\documentstyle[aclap]{article}

\title{\vspace{-0.5in}Using textual clues to improve metaphor processing}
\author{St\'ephane Ferrari\\
  LIMSI-CNRS\\
  PO Box 133\\
  F-91403 Orsay c\'edex, {\sc France}\\
  ferrari@limsi.fr}

\begin{document}

\maketitle
\vspace{-0.5in}
\begin{abstract}
  In this paper, we propose a textual clue approach to help metaphor
  detection, in order to improve the semantic processing of this
  figure. The previous works in the domain studied the semantic
  regularities only, overlooking an obvious set of regularities. A
  corpus-based analysis shows the existence of surface regularities
  related to metaphors. These clues can be characterized by syntactic
  structures and lexical markers. We present an object oriented model
  for representing the textual clues that were found. This
  representation is designed to help the choice of a semantic
  processing, in terms of possible non-literal meanings. A prototype
  implementing this model is currently under development, within an
  incremental approach allowing step-by-step evaluations.
  \footnote{This work takes part in a research project sponsored by
    the AUPELF-UREF (Francophone Agency For Education and Research)}
\end{abstract}

\section{Introduction}
\label{sec:intro}

Metaphor is a frequently used figure of speech, reflecting common
cognitive processes. Most of the previous works in Natural Language
Understanding (NLU) looked for regularities only on the semantic side
of this figure, as shown in a brief overview in section
\ref{sec:overview}. This resulted in complex semantic processings, not
based on any previous robust detection, or requiring large and
exhaustive knowledge bases. Our aim is to provide NLU systems with a
set of heuristics for choosing the most adequate semantic processing,
as well as to give some probabilistic clues for disambiguating the
possibly multiple meaning representations.

A corpus-based analysis we made showed the existence of textual clues
in relation with the metaphors. These clues, mostly lexical markers
combined with syntactic structures, are easy to spot, and can provide
a first set of detection heuristics. We propose, in section
\ref{sec:clues}, an object oriented model for representing these clues
and their properties, in order to integrate them in a NLU system. For
each class, attributes give information for spoting the clues, and,
when possible, the source and the target of the metaphor, using the
results of a syntactic parsing. A prototype, STK, partially
implementing the model, is currently under development, within an
incremental approach. It is already used to evaluate the clues
relevance.

In conclusion, we will discuss how the model can help chosing the
adequate semantic analysis to process at the sentence level or
disambiguating multiple meaning representations, providing
probabilities for non-literal meanings.

\section{Classical methods: a brief overview}
\label{sec:overview}

The classical NLU points of view of metaphor have pointed out the
multiple kinds of relations between what is called the source and the
target of the metaphor, but rarely discuss the problem of detecting
the figure that bears the metaphor. For our purpose, we choose to
present these approaches in two main groups, depending on how they
initiate the semantic processing.

The previous works led to a classification introduced by Dan Fass
\cite{Fas91}. In the {\it comparison view}, the metaphor corresponds
to an analogy between the structures representing the source and the
target of the figure, as in Gentner's works \cite{Gen88a} and their
implementation \cite{Fal89}. The {\it interaction view}, as in Hobbs
\cite{Hob91}, points at the novelty brought by the metaphor. Fass also
distinguishes a {\it selection restrictions violations view}
presenting the metaphor as a kind of anomaly. We would argue that the
two previous views already considered metaphor as a kind of anomaly.
Indeed, the semantic analysis proposed for dealing with metaphors were
processed depending on the results of another, say a ``classical''
one\footnote{We prefer to call it a {\it classical} rather than {\it
    literal meanings} processing because it can deal with some
  conventional metaphors, even if not explicitly mentioned.}. Thereby,
detecting a metaphor meant detecting an anomaly in the meaning
representation issued from such a classical analysis.

Fass proposed a method for discriminating literal meanings, metaphors,
metonymies and ``anomalies'', merging different points of view
\cite{Fas91}. In this approach, multiple semantic analysis can be
processed, resulting in possibly multiple meaning representations. In
\cite{Pri92b}, a method to overcome similar kinds of ambiguities
reveal the difficulties encountered if no previous detection is made.
James Martin's approach \cite{Mar92}, called the {\it conventional
  view} by Fass, is based on Lakoff's theory on cognitive metaphors
\cite{Lak80}.  It requires a specific knowledge representation base
and also results in multiple representation meanings. Detecting a
metaphor is meaningless here, and conventional metaphoric meanings can
be viewed as polysemies. Martin revealed at least that the heuristic
of the ill-formness of meaning representations issued from classical
analysis is not sufficient at all to deal with all the possible
metaphors.

In our point of view, all the previous approaches were founded. The
main remaining problem, however, is to choose an adequate processing
when confronted with a metaphor, and thus, to detect the metaphors
before trying to build their meaning representation. This can be
partially solved using textual clues.

\section{Textual clues: object oriented description}
\label{sec:clues}

If the classical views of the metaphor overlook the textual clues, in
other domains, especially those concerning explanation, they have been
wisely re-introduced. In \cite{Per90}, Pery-Woodley shows the
existence of such clues related to the explanatory discourse. They can
help in generating explanations in natural language as well as in
modelling the student in a intelligent tutoring system \cite{Dan92}. A
corpus of 26 explanatory texts in French, of about 200 words each, has
been collected under a shared research project between psychologists
and computer scientists, in order to study metaphors and analogies in
teaching. The analysis we made showed the existence of textual clues
in relation with metaphoric contexts and analogies (e.g. ``like'',
``such as'', ``illustrated by''). They can be characterized by
syntactic regularities (e.g. the comparative is used in structures
such as ``less than'', ``more than''; the identification is made
through attributes or appositions, ...). They also involve lexical
markers (e.g.  ``literaly'', ``illustrating'', ``metaphorically'',).
These properties, already found in the previous works, can help
detecting the clues themselves.  Studying the relation between the
syntactic regularities and the lexical markers, one can observe that
the first build the ground where to find the second. We thus propose
an object-oriented model for representing these clues. A generic
textual clue can thereby be described by the two following attributes:
\begin{itemize}
\item the {\it Surface Syntactic Pattern} representing the syntactic
  regularity, with a label on the item where to find the lexical
  marker
\item the {\it Lexical Marker} itself
\end{itemize}

Typically, the word ``metaphor'' itself can be used as a lexical
marker in expressions such as {\it ``to extend the conventional
  metaphor, pruning such a tree means to generalize''}. On the other
hand, ``metaphor'' will not be a marker if used as the subject of the
sentence, like in this one. Thus, describing the syntactic
regularities surrounding a lexical marker improves its relevance as a
marker. We propose to represent this relevance for probabilistic
purposes. Each clue that was found is currently evaluated on a large
corpus (about 450,000 words). The frequencies of use of the lexical
markers in metaphoric contexts are represented in the relevance
attribute (see example below).

The syntactic structures may also give information about the source
and the target of the metaphor. For instance, in the sentence {\it
  ``Yesterday, at home, Peter threw himself on the dessert like a
  lion.''}, the subject inherits the properties of speed and voracity
of a lion attacking its victim. It is here possible to spot the source
and the target of the metaphor using the syntactic properties of the
comparison. Two attributes are added to textual clues related to
metaphors, corresponding to the elements of the sentence bearing the
source and the target.

\vspace*{2mm}
\noindent {\bf Example of textual clue representations}
\begin{description}
\itemsep=0pt
\parsep=0pt
\parskip=0pt
\topsep=0pt
\partopsep=0pt
\item[type] metaphor-analogy
\item[name] B.2.2.2
\item[comment] {\it comparison involving the meaning of a marker,
    adjective, attribute of the object, object before the verb}
\item[SSP] $GN_0 \quad GN_1 \quad V_1 \quad
  Adj_0 \quad [prep] \quad GN_2$
\item[LM] $Adj_0$: pareil {\it (meaning ``similar'')}
\item[target] $GN_1$
\item[source] $GN_2$
\item[LM relevance] (15/28)
  \begin{description}
\vspace*{-1mm}
  \item[number of occurrences] 28
\vspace*{-1mm}
  \item[conventional metaphors] 3
\vspace*{-1mm}
  \item[new metaphors] 2
\vspace*{-1mm}
  \item[metaphoric contexts] 12
\vspace*{-1mm}
  \item[total] 15
  \end{description}
\end{description}
\vspace*{-2mm}

{\small {\bf Notations}: $GN$ and $GV$ stand for nominal or verbal
  groups, $Adj$ and $Adv$ for adjectives and adverbs, and $prep$ for
  prepositions.}
\vspace*{2mm}

The model has been partially implemented in a tool, STK, for detecting
the textual clues related to metaphors and adding specific marks when
found. In its current version, STK allows us to tokenize, tag, and
search for lexical markers on large corpora. The tagger we use is the
one developped by Eric Brill \cite{Bri92} with a set of tags
indicating the grammatical categories as well as other information
such as the number and the gender for nouns and adjectives. It is
evaluated under GRACE\footnote{GRACE stands for ``Grammars and
  Resources for Corpora Analysis and their Evaluation''. It is a
  national research project for the development of tools for French
  language processing.} protocol for corpus-oriented tools assigning
grammatical categories.  It is currently used for the evaluation of
the textual clues that were found. The latter can be easily retrieved
using STK, avoiding lexical ambiguities. They are then analyzed by
hand, in order to determine their relevance attribute. In the previous
example of textual clue, the relevance values are issued from this
corpus-based analysis.

\section{Conclusion, perspectives}

Classical approaches to the metaphor in NLU revealed multiple
underlying processes. We therefore focussed our study on how to help
detecting metaphors in order to chose the most adequate semantic
processing. Textual clues can give information about the figures that
bear the metaphor, which are easy to spot. Indeed, they can be found
using the results of syntactic parsing. We proposed an object-oriented
model to represent these clues and their multiple properties.

If textual clues give information about possible non-literal meanings,
metaphors and analogies, one may argue they do not allow for a robust
detection. Indeed, a textual clue is not sufficient to prove the
presence of such figures of speech. The relevance of each clue can be
used to help disambiguating multiple meaning representation when it
occurs. This must not be the only disambiguation tool, but when no
other is avalaible, it provides NLU systems with a probabilistic
method.

Our future works will focuss on the study of the relation between the
metaphors introduced by a clue and others that are not conventional.
The guideline is that novel metaphors not introduced by a clue at the
sentence level may have been introduced previously in the text.

\bibliographystyle{acl}

\end{document}